\documentclass[10pt,twocolumn,letterpaper]{article}

\usepackage{titling}
\usepackage{wacv}
\usepackage{times}
\usepackage{epsfig}
\usepackage{graphicx}
\usepackage{amsmath}
\usepackage{amssymb}
\usepackage{booktabs}
\usepackage{makecell}
\usepackage{multirow}
\usepackage{tipa}
\usepackage{adjustbox}
\def\wacvPaperID{1091} % Enter the WACV Paper ID here

\wacvalgorithmstrack   % Uncomment this line if you are submitting to the Algorithms Track.
\wacvfinalcopy % *** Uncomment this line for the final submission

\ifwacvfinal
\usepackage[breaklinks=true,bookmarks=false]{hyperref}
\else
\usepackage[pagebackref=true,breaklinks=true,colorlinks,bookmarks=false]{hyperref}
\fi

\begin{document}

%%%%%%%%% TITLE
\title{VEATIC: Video-based Emotion and Affect Tracking in Context Dataset}

% \author{Zhihang Ren*\quad Jefferson Ortega*\quad Yifan Wang*\quad Zhimin Chen\quad David Whitney\\
% University of California, Berkeley\\
% {\tt\small \{peter.zhren,jefferson\_ortega,wyf020803,zhimin,dwhitney\}@berkeley.edu}
% \and
% Yunhui Guo\\
% University of Texas at Dallas\\
% {\tt\small yunhui.guo@utdallas.edu}
% \and
% Stella X. Yu\\
% University of Michigan, Ann Arbor\\
% {\tt\small stellayu@umich.edu}
% }

\author{Zhihang Ren*$^{1}$, Jefferson Ortega*$^{1}$, Yifan Wang*$^{1}$, Zhimin Chen$^{1}$, Yunhui Guo$^{2}$,\\ Stella X. Yu$^{1,3}$, David Whitney$^{1}$\\
$^{1}$University of California, Berkeley, $^{2}$University of Texas at Dallas,\\ $^{3}$University of Michigan, Ann Arbor\\
$^{1}${\tt\small \{peter.zhren,jefferson\_ortega,wyf020803,zhimin,dwhitney\}@berkeley.edu},\\
$^{2}${\tt\small yunhui.guo@utdallas.edu},$^{3}${\tt\small stellayu@umich.edu}
}

\date{}
\maketitle
\thispagestyle{empty}

\def\thefootnote{*}\footnotetext{These authors contributed equally to this work.}

%%%%%%%%% ABSTRACT
\begin{abstract}
   Human affect recognition has been a significant topic in psychophysics and computer vision. However, the currently published datasets have many limitations. For example, most datasets contain frames that contain only information about facial expressions. Due to the limitations of previous datasets, it is very hard to either understand the mechanisms for affect recognition of humans or generalize well on common cases for computer vision models trained on those datasets. In this work, we introduce a brand new large dataset, the Video-based Emotion and Affect Tracking in Context Dataset (\textbf{VEATIC}), that can conquer the limitations of the previous datasets. VEATIC has $124$ video clips from Hollywood movies, documentaries, and home videos with continuous valence and arousal ratings of each frame via real-time annotation. Along with the dataset, we propose a new computer vision task to infer the affect of the selected character via both context and character information in each video frame. Additionally, we propose a simple model to benchmark this new computer vision task. We also compare the performance of the pretrained model using our dataset with other similar datasets. Experiments show the competing results of our pretrained model via VEATIC, indicating the generalizability of VEATIC. Our dataset is available at https://veatic.github.io.
\end{abstract}
%%%%%%%%% BODY TEXT
\section{Introduction}

Recognizing human affect is of vital importance in our daily life. We can infer people's feelings and predict their subsequent reactions based on their facial expressions, interactions with other people, and the context of the scene. It is an invaluable part of our communication. Thus, many studies are devoted to understanding the mechanism of affect recognition. With the emergence of Artificial Intelligence (AI), many studies have also proposed algorithms to automatically perceive and interpret human affect, with the potential implication that systems like robots and virtual humans may interact with people in a naturalistic way.

\begin{figure}[!th]
    \centering
    \includegraphics[width=0.4\textwidth]{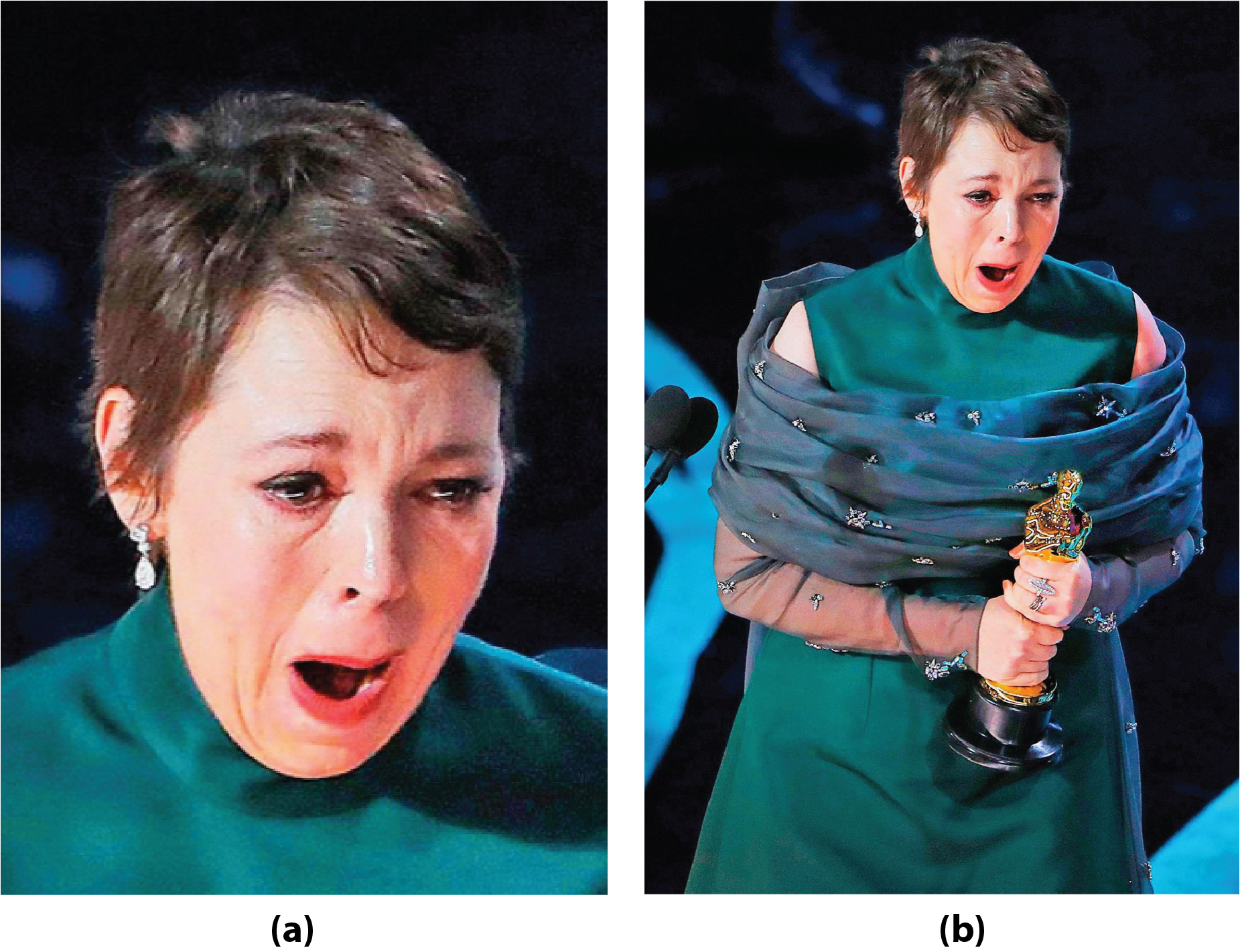}
    \caption{Importance of context in emotion recognition. How does she feel? Look at the woman in picture (a). If you had to guess her emotion, you might say that she is sad or in grief. However, picture (b) reveals the context of the scene allowing us to correctly observe that she is very happy or excited.}
    \label{fig:concept}
\end{figure}

%How human perceive emotion
When tasked with emotion recognition in the real world, humans have access to much more information than just facial expressions. Despite this, many studies that investigate emotion recognition often use static stimuli of facial expressions that are isolated from context, especially in assessments of psychological disorders~\cite{baron2001reading,gao2021facial} and in computer vision models~\cite{tarnowski2017emotion,tumen2017facial}. Additionally, while previous studies continue to investigate the process by which humans perceive emotion, many of these studies fail to probe how emotion recognition is influenced by contextual factors like the visual scene, background information, body movements, other faces, and even our beliefs, desires, and conceptual processing ~\cite{barrett2010context,lee2012context,chen2019tracking,nook2015new,ong2019computational}. Interestingly, visual contextual information has been found to be automatically and effortlessly integrated with facial expressions~\cite{aviezer2011automaticity}. It can also override facial cues during emotional judgments~\cite{kayyal2015context}(Figure~\ref{fig:concept}), and can even influence emotion perception at the early stages of visual processing~\cite{calbi2019context}. In fact, contextual information is often just as valuable to understand a person's emotion as the face itself~\cite{chen2019tracking,chen2021inferential,chen2022inferential}. The growing evidence of the importance of contextual information in emotion recognition~\cite{barrett2010context} demands that researchers reevaluate the experimental paradigms in which they investigate human emotion recognition. For example, to better understand the mechanisms and processes that lead to human emotion recognition during everyday social interactions, the generalizability of research studies should be seriously considered. Most importantly, datasets for emotion and affect tracking should not only contain faces or isolated specific characters, but contextual factors such as background visual scene information, and interactions between characters should also be included.

In order to represent the emotional state of humans, numerous studies in Psychology and Neuroscience have proposed methods to quantify humans' emotional state which include both categorical and continuous models of emotion. The most famous and dominant categorical theory of emotion is the theory of basic emotions which states that certain emotions are universally recognized across cultures (anger, fear, happiness, etc.) and that all emotions differ in their behavioral and physiological response, their appraisal, and in expression~\cite{ekman1992argument}. Alternatively, the circumplex model of affect, a continuous model of emotion, proposes that all affective states arise from two neurophysiological systems related to valence and arousal and all emotions can be described by a linear combination of these two dimensions ~\cite{russell1980circumplex,posner2005circumplex,russell1997dimensional}. Another model of emotion recognition, the Facial Action Coding System model, states that all facial expressions can be broken down into the core components of muscle movements called Action Units~\cite{ekman1978facial}. Previous emotion recognition models have been built with these different models in mind~\cite{tian2001recognizing,valenza2011role,nasri2020face}. However, few models focus on measuring affect using continuous dimensions, an unfortunate product of the dearth of annotated databases available for affective computing.

%Existing datasets analysis
Based on the aforementioned emotion metrics, many emotion recognition datasets have been developed. Early datasets, such as SAL~\cite{douglas2008sensitive}, SEMAINE~\cite{mckeown2011semaine}, Belfast induced~\cite{sneddon2011belfast}, DEAP~\cite{koelstra2011deap}, and MAHNOB-HCI~\cite{soleymani2011multimodal} are collected under highly controlled lab settings and are usually small in data size. These previous datasets lack diversity in terms of characters, motions, scene illumination, and backgrounds. Moreover, the representations in early datasets are usually discrete. Recent datasets, like RECOLA~\cite{ringeval2013introducing}, MELD~\cite{poria2018meld}, OMG-emotion dataset~\cite{barros2018omg}, Aff-Wild~\cite{zafeiriou2017aff}, and Aff-Wild2~\cite{kollias2018aff,kollias2019expression}, start to collect emotional states via continuous ratings and utilize videos on the internet or called "in-the-wild". However, these datasets lack contextual information and focus solely on facial expressions. The frames are dominated by characters or particular faces. Furthermore, the aforementioned datasets have limited annotators (usually less than 10). As human observers have strong individual differences and suffer from many biases~\cite{davidoff2012differences,partos2016you,ren2023serial}, limited annotators can lead to substantial annotation biases.

%Our datasets
In this study, we introduce the Video-based Emotion and Affect Tracking in Context Dataset (\textbf{VEATIC}, /ve\textprimstress\ae t\textsci c/), a large dataset that can be beneficial to both Psychology and computer vision groups. The dataset includes $124$ video clips from Hollywood movies, documentaries, and home videos with continuous valence and arousal ratings of each frame via real-time annotation. We also recruited a large number of participants to annotate the data. Based on this dataset, we propose a new computer vision task, i.e., automatically inferring the affect of the selected character via both context and character information in each video frame. In this study, we also provide a simple solution to this task. Experiments show the effectiveness of the method as well as the benefits of the proposed VEATIC dataset. In a nutshell, the main contributions of this work are:

\begin{itemize}
    \item We build the first large video dataset, \textbf{VEATIC}, for emotion and affect tracking that contains both facial features and contextual factors. The dataset has continuous valence and arousal ratings for each frame.
    \item In order to alleviate the biases from annotators, we recruited a large set of annotators ($192$ in total) to annotate the dataset compared to previous datasets (usually less than 10).
    \item We provide a baseline model to predict the arousal and valence of the selected character from each frame using both character information and contextual factors.
\end{itemize}

\begin{table*}[th]
\footnotesize
\centering
\begin{tabular}{|c|c|c|c|c|c|c|c|}
\hline
\textbf{Database} 
& \textbf{Annotation Type} 
& \textbf{Condition}
& \textbf{\makecell[c]{\# videos}} 
& \textbf{\makecell[c]{Length of\\ Videos}} 
& \textbf{\makecell[c]{\# Annotators}}
& \textbf{Context}\\ \hline

SAL~\cite{douglas2008sensitive} & \makecell[c]{Valence-Arousal\\(Continuous)} & Controlled & 23 & \makecell[c]{SAL 0: 5min \\ SAL 1: 30min} & 4 & $\times$ \\ \hline

SEMAINE~\cite{mckeown2011semaine} &  Mixed* & Controlled & $\backslash$ & Total: 6.5hours & 6-8 & $\times$ \\ \hline

SEND~\cite{ong2019modeling} & valence & Controlled & 193 & 135s & 700 & $\times$ \\ \hline

Belfast induced~\cite{sneddon2011belfast} & Mixed & Controlled & 37 & 5-60s & 6-258 & $\times$ \\ \hline

MAHNOB-HCI~\cite{soleymani2011multimodal} & Mixed & Controlled & 20 & 34.9-117s & 50 & $\times$ \\ \hline

MELD~\cite{poria2018meld} & \makecell[c]{7 Emotion\\Categories} & In-the-Wild & 1,433 & 3.59s & 3 & $\times$ \\ \hline

OMG Emotion~\cite{barros2018omg} & Mixed & In-the-Wild & 567 & 1min & 5 & $\times$ \\ \hline

RECOLA~\cite{ringeval2013introducing} & \makecell[c]{Valence-Arousal\\(Continuous)} & Controlled & 46 & 5min & 6 & $\times$ \\ \hline

AFEW~\cite{dhall2012collecting} & \makecell[c]{7 Basic Facial \\Expression} & In-the-Wild & 1,809 & 0.3s - 5.4s & 3 & $\times$ \\ \hline

AFEW-VA~\cite{kossaifi2017afew} & \makecell[c]{Valence-Arousal\\(Discrete)} & In-the-Wild & 600 & 0.5s - 4s & 2 & $\times$ \\ \hline

Aff-Wild~\cite{zafeiriou2017aff} & \makecell[c]{Valence-Arousal\\(Continuous)} & In-the-Wild & 298 & 6s - 14min28s & 8 & $\times$ \\ \hline

Aff-Wild2~\cite{kollias2018aff,kollias2019expression} & \makecell[c]{Valence-Arousal\\(Continuous)} & In-the-Wild & 260 & 4s - 15min4s & 4 & $\times$ \\ \hline

AM-FED~\cite{mcduff2013affectiva} & 12 Action Units & In-the-Wild & 242 &  49.69s & $\backslash$ & $\times$ \\ \hline

DEAP~\cite{koelstra2011deap} & \makecell[c]{Valence-Arousal\\-Dominance\\(Discrete)} & \makecell[c]{Music\\Videos} & 120 & 1min & 14-16 & \textcolor{red}{\checkmark} \\ \hline

CAER~\cite{lee2019context} & \makecell[c]{7 Emotion\\Categories} & In-the-Wild & 13,201 & 1s - 5s & 6 & \textcolor{red}{\checkmark} \\ \hline

CAER-S~\cite{lee2019context} & \makecell[c]{7 Emotion\\Categories} & \makecell[c]{In-the-Wild\\Image-based} & $\backslash$ & 70,000 images & 6 & \textcolor{red}{\checkmark} \\ \hline

EMOTIC~\cite{kosti2019context} & Mixed & \makecell[c]{In-the-Wild\\Image-based} & $\backslash$ & 18,316 images & 3-5 & \textcolor{red}{\checkmark} \\ \hline

\textbf{\textit{VEATIC}} (ours) & \makecell[c]{\textcolor{red}{Valence-Arousal}\\\textcolor{red}{(Continuous)}} & \textcolor{red}{In-the-Wild} & \textcolor{red}{124} & \textcolor{red}{10s - 2min37s} &  \textcolor{red}{192} & \textcolor{red}{\checkmark} \\ \hline
\end{tabular}
\newline
\caption{Comparison of the VEATIC dataset with existing emotion recognition datasets. VEATIC contains a large amount of video clips and a long video total duration. It is the first large context-aware emotion recognition video dataset with continuous valence and arousal annotations. VEATIC also has many more annotators compared to other context-aware emotion recognition video datasets. (*: Mixed means containing both continuous and categorical annotations.)}
\label{table:comparisonD}
\end{table*}

\vspace{-0.2cm}
\section{Related Work}
%most related datasets
Recently, there have been several datasets that provide frames with both facial and context information, like CAER~\cite{lee2019context} and EMOTIC~\cite{kosti2019context}. CAER~\cite{lee2019context} is a video-based dataset that contains categorical labels of each video frame, and EMOTIC~\cite{kosti2019context} is an image-based dataset containing both categorical expression labels and continuous valence-arousal-dominance ratings. Unlike these datasets, our dataset is video-based and contains continuous valence and arousal ratings. A detailed comparison between our dataset with previous datasets can be found in Table~\ref{table:comparisonD}.

%Computer Vision models
Based on various emotion datasets, studies have started to focus on how to infer emotion automatically. Human affect can be inferred from many modalities, such as audio~\cite{zhang2018attention,yoon2018multimodal,venkataramanan2019emotion}, visual~\cite{mittal2020emoticon,savchenko2021facial,savchenko2022classifying,luo2022learning}, and text~\cite{yoon2018multimodal,hipson2021emotion}. For visual inputs, in particular, there are three major tasks. The valence-arousal estimation task aims to predict the valence and arousal of each image/frame~\cite{zhang2020m,zafeiriou2017aff,kollias2018aff,kollias2019expression}; the expression recognition task focuses on classifying emotional categories of each image/frame~\cite{wang2020suppressing,she2021dive,xue2022coarse}; and the action unit (AU) detection task intends to detect facial muscle actions from the faces of each image/frame~\cite{kaili2016deep,shao2018deep,li2019self,vemulapalli2019compact}. Currently, most proposed methods rely highly on the facial area to infer the emotional state. Indeed, the facial area contains rich information about the human emotional state. However, contextual factors also provide essential information that is necessary for humans to correctly infer and perceive the emotional states of others ~\cite{chen2019tracking,chen2021inferential,chen2022inferential}. Several studies~\cite{lee2019context,kosti2019context,mittal2020emoticon} have started to incorporate context information as a source of affect inference. In this study, we also adopted both facial and context information to achieve the new task, i.e., to infer the valence and arousal for each video frame.

To infer the affect of a person, we usually need to deal with temporal information of either audio segments, video frames, or words. Many studies~\cite{yoon2018multimodal,zafeiriou2017aff,kollias2018aff,kollias2019expression} started to utilize long short term memory (LSTM)~\cite{hochreiter1997long}, gated recurrent unit (GRU)~\cite{cho2014learning}, or recurrent neural network (RNN)~\cite{hopfield1982neural,rumelhart1985learning} to process the temporal information. With the emergence of the visual transformer (ViT)~\cite{dosovitskiy2020image}, attention has been shifted. Many video understanding tasks~\cite{girdhar2019video,arnab2021vivit,liu2022video} have utilized ViT for temporal information understanding and achieving state-of-the-art performance. Our baseline method also adopted ViT as a tool to process the temporal information in video clips.

\begin{figure*}[th]
    \centering
    \includegraphics[width=0.85\textwidth]{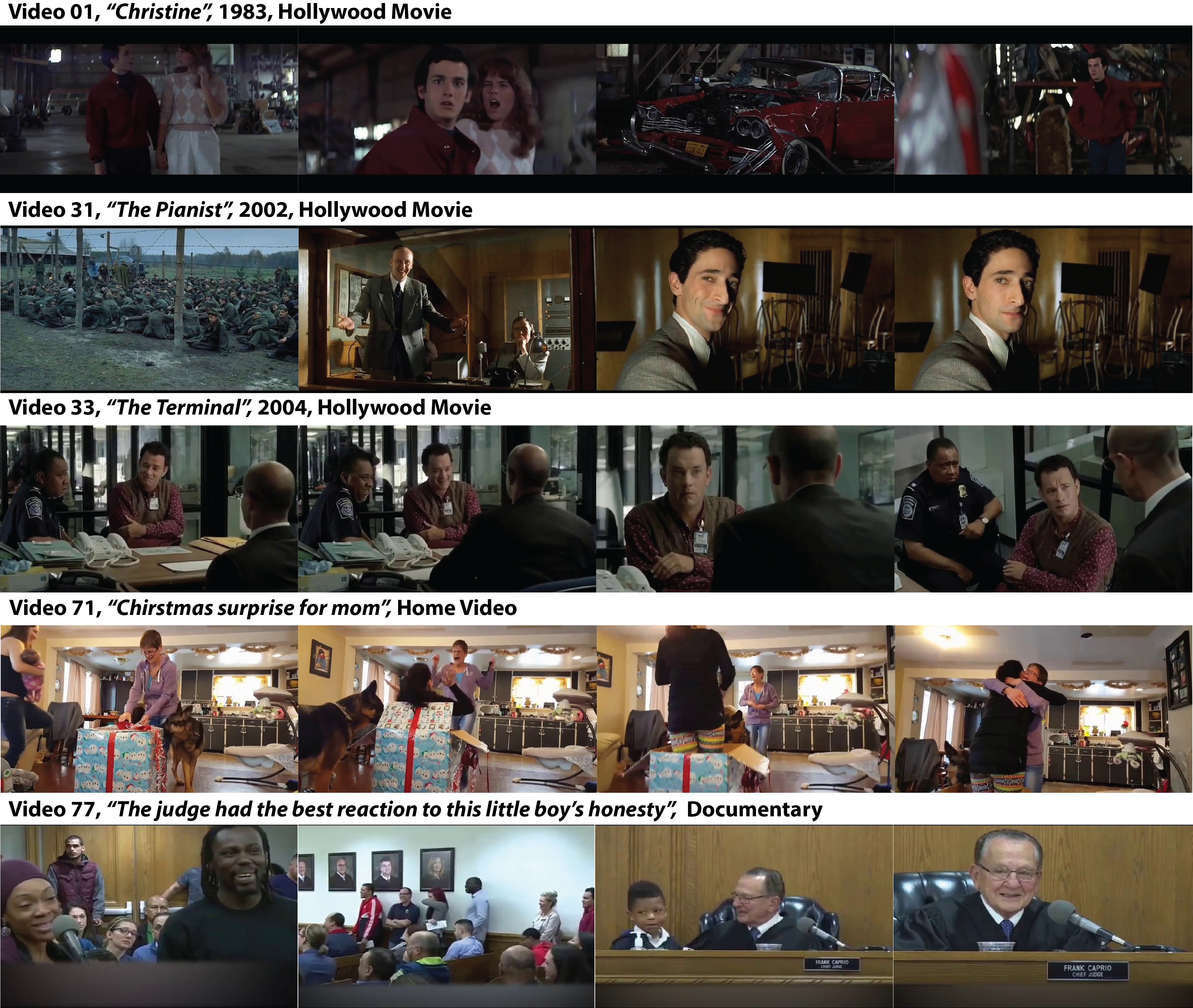}
    \caption{Overview of video frames in VEATIC. We sampled 4 key frames from 5 videos in our dataset. Unlike other datasets where the source of video clips is unique, video clips of VEATIC come from different sources. They include Hollywood movies, documentaries, and homemade videos.  Thus, it would make the model trained on our dataset have more generalization ability. For the visual input, VEATIC contains various context information, including different backgrounds, lighting conditions, character interactions, etc. It makes the dataset more representative of our daily life. At last, the emotion/affect of the selected character varies a lot in each video clip, making modeling the character's affect in VEATIC more challenging.}
    \label{fig:preview}
\end{figure*}

\vspace{-0.1cm}
\section{VEATIC Dataset}
In this section, we introduce the Video-based Emotion and Affect Tracking in Context Dataset (\textbf{VEATIC}). First, we describe how we obtained all the video clips. Next, we illustrate the data annotation procedures and pre-processing process. Finally, we report important dataset statistics and visualize data analysis results.

\subsection{Video Clips Acquisition}
All video clips used in the dataset were acquired from an online video-sharing website (YouTube) and video clips were selected on the basis that the emotions/affect of the characters in the clips should vary across time. In total, the VEATIC dataset contains 124 video clips, 104 clips from Hollywood movies, 15 clips from home videos, and 5 clips from documentaries or reality TV shows. Sample frames from the VEATIC dataset are shown in (Figure~\ref{fig:preview}). These videos contain zero to multiple interacting characters. All sound was removed from the videos so observers only had access to visual information when tracking the emotion of the target character.

\begin{figure*}[th]
    \centering
    \includegraphics[width=0.9\textwidth]{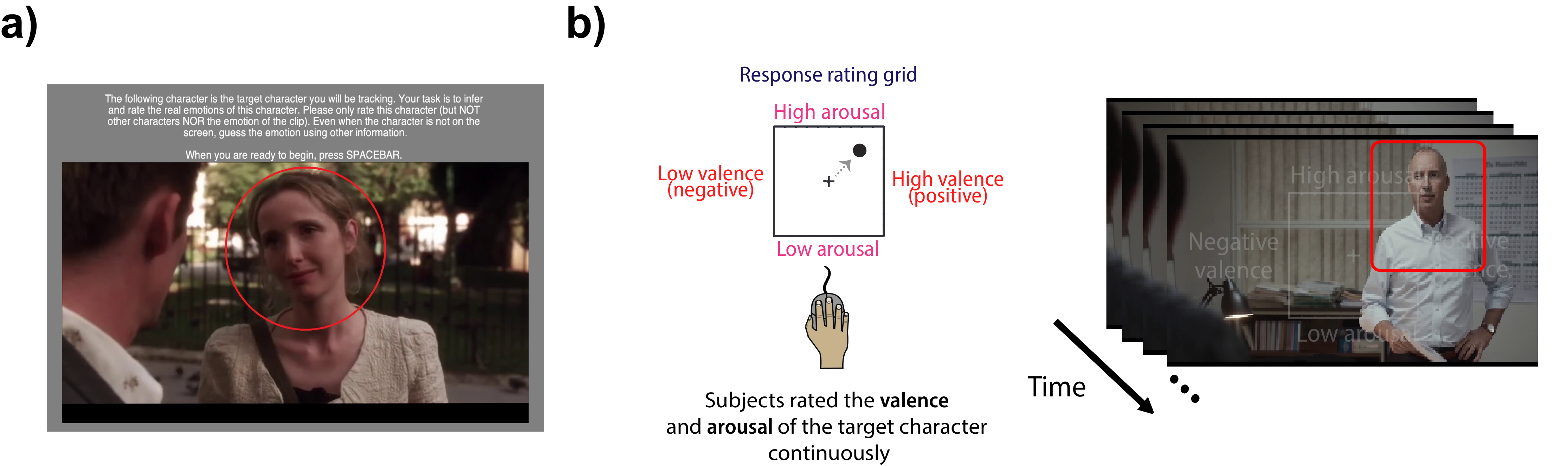}
    \caption{User interface used for video annotation. a) Participants were first shown the target character and were reminded of the task instructions before the start of each video. b) The overlayed valence and arousal grid that was present while observers annotated the videos. Observers were instructed to continuously rate the emotion of the target character in the video in real-time. If observers did not move their mouse for more than 10 seconds, the response rating grid would flash to remind the observer to continuously rate the emotion.}
    \label{fig:task}
\end{figure*}

\subsection{Data Annotation and Pre-processing}
In total, we had 192 observers who participated in the annotation of the videos in the dataset. All participants provided signed consent in accordance with the guidelines and regulations of the UC Berkeley Institutional Review Board and all experimental procedures were approved.

Participants watched and rated a total of 124 videos in the dataset. To prevent observers from getting fatigued, we split the annotation procedure into two 1-hour and 30-minute annotation sessions. Before participants were able to annotate any videos, they were shown a printed version of the valence-arousal affect rating grid with example emotions labeled in different locations of the grid according to the ratings provided by Bradley and Lang (1999)~\cite{bradley1999affective}. Annotators were instructed to familiarize themselves with the dimensions and the sample word locations which they would later utilize in the annotation process. After participants familiarized themselves with the affect rating grid, they then completed a two-minute practice annotation where they continuously tracked the valence and arousal of a target character in a video (Figure~\ref{fig:task}b). Annotators were instructed to track the valence and arousal of the target character in the video by continuously moving their mouse pointer in real-time within the 2D valence-arousal grid. The grid would map to their valence and arousal ratings in the range of $[-1, 1]$. To control for potential motor biases, we counterbalanced the valence-arousal dimensions between participants where half of the annotators had valence on the x-axis and arousal on the y-axis and the other half had the dimensions flipped so that arousal was on the x-axis and valence was on the y-axis. Once observers finished the practice annotation session, they then started annotating the videos in the dataset. 

Before participants started the annotations, they were shown an image with the target character circled (Figure~\ref{fig:task}a) which informs the participants which character they will track when the video begins. Then, they annotated the video clips in real-time. At the end of each video annotation, participants reported their familiarity with the video clip using a 1-5 discrete Likert scale that ranged from "Not familiar", "Slightly familiar", "Somewhat familiar", "Moderately familiar", and "Extremely familiar". Participants were also asked about their level of enjoyment while watching the clip which was rated using a 1-9 discrete Likert scale that ranged from 1 (Not Enjoyable) to 9 (Extremely Enjoyable). Additionally, in order to not make participants feel bored, all 124 video clips were split into two sessions. Participants rated the video clips in two sessions separately.

During each trial, we assessed whether participants were not paying attention by tracking the duration that they kept the mouse pointer at any single location. If the duration was longer than 10 seconds, the affect rating grid would start to fluctuate which reminded participants to continue tracking the emotion of the target character. In order to assess whether there were any noisy annotators in our dataset, we computed each individual annotator's agreement with the consensus by calculating the Pearson correlation between each annotator and the leave-one-out consensus (aggregate of responses except for the current annotator) for each video. We found that only one annotator had a correlation lower than .2 across all videos with the leave-one-out consensus. Since only one annotator fell below our threshold, we decided to keep the annotator in the dataset in order to not remove any important alternative annotations to the videos.

\subsection{Visualization and Data Analysis}
\begin{figure}[t]
    \centering
    \includegraphics[width=0.45\textwidth]{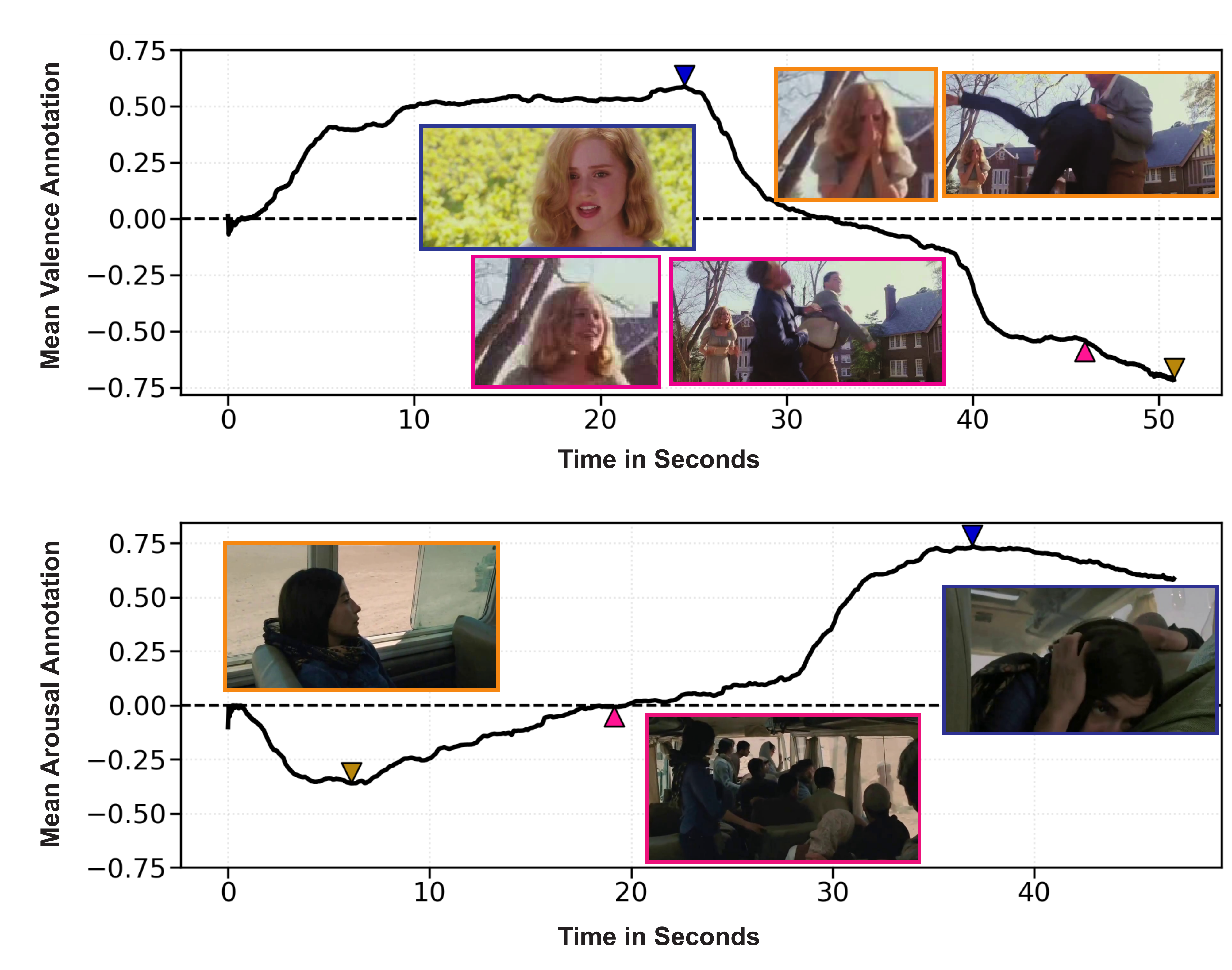}
    \caption{Visualization of sample mean ratings of valence and arousal for specific video clips with the zoom-in view of the selected character. We show key frames related to specific mean ratings of valence and arousal. Corresponding frames and ratings are marked the same color.}
    \label{fig:ratingvisual}
\end{figure}

\begin{figure}[h]
    \centering
    \includegraphics[width=0.45\textwidth]{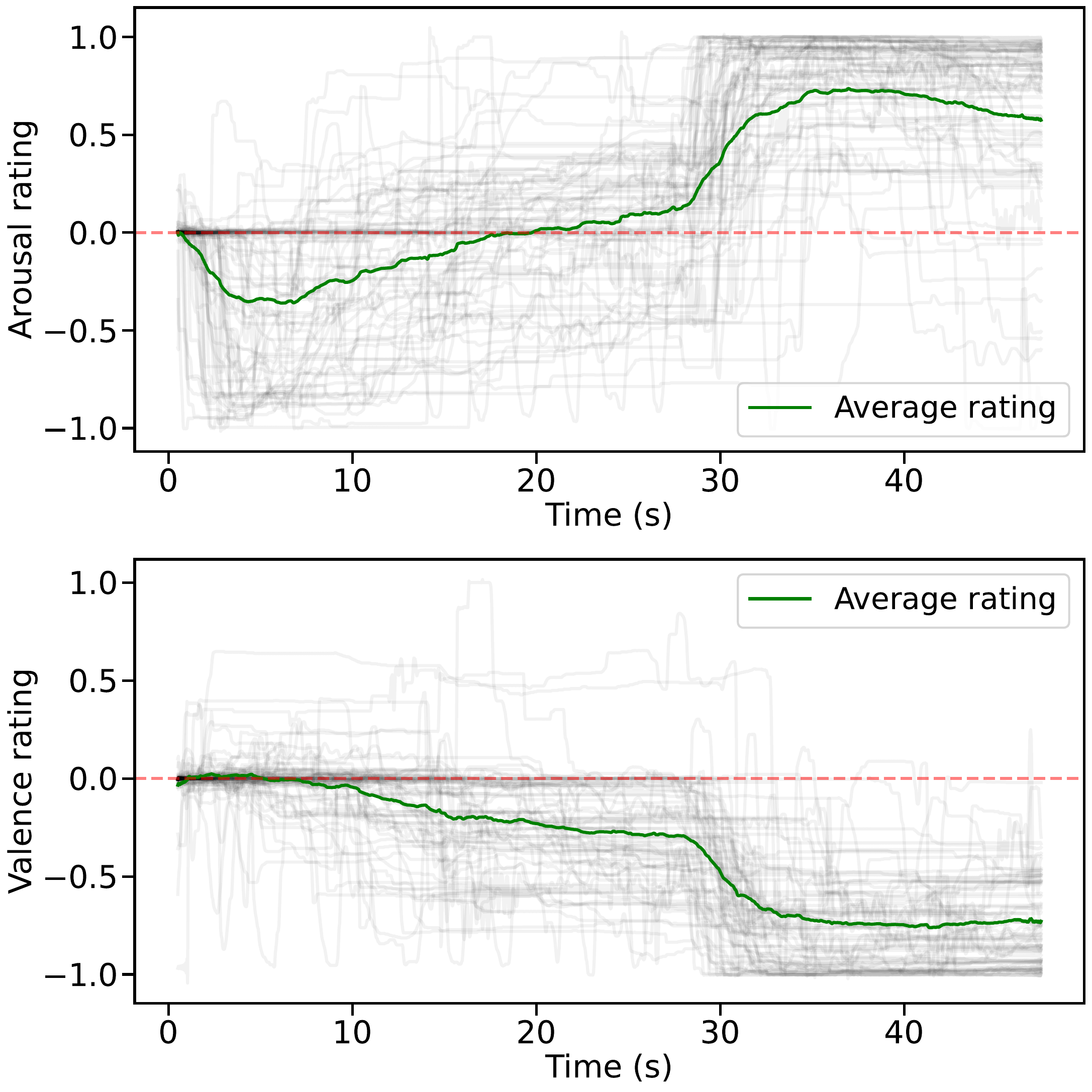}
    \caption{Example valence and arousal ratings for a single video (video 47). Transparent gray lines indicate individual subject ratings and the green line is the average rating across participants.}
    \label{fig:example_rating}
\end{figure}

Figure~\ref{fig:ratingvisual} shows sample mean ratings and key frames in 2 different video clips. Clearly, both the valence and arousal here have a wide range of ratings. Moreover, it shows that context information, either spatial and/or temporal, plays an important role in emotion recognition tasks. In the valence example (upper figure), without the temporal and/or spatial context information of the fighting, it would be hard to recognize whether the character (the woman) in the last frame (yellow) is surprisingly happy or astonished. In the arousal example (lower figure), even without the selected character's face, observers can easily and consistently infer the character's arousal via the intense context.

Figure~\ref{fig:example_rating} illustrates sample valence and arousal ratings of all participants for a single video in our dataset. Individual subject's ratings (gray lines) followed the consensus ratings across participants (green line) for both valence and arousal ratings. The dense gray line overlapping around the green consensus line indicates agreements between a wide range of observers. Additionally, We investigated how observers' responses varied across videos by calculating the standard deviation across observers for each video. 
% (shown in Figure~\ref{fig:agreement})
 We found that the variance between observers for both valence and arousal dimensions was small with valence having an average standard deviation of $\mu=0.248$ and a median of $0.222$ and arousal having an average standard deviation of $\mu=0.248$ and a median of $0.244$, which are comparable with the valence and arousal rating variance from EMOTIC~\cite{kosti2019context}.

\begin{figure}[th]
    \centering
    \includegraphics[width=0.45\textwidth]{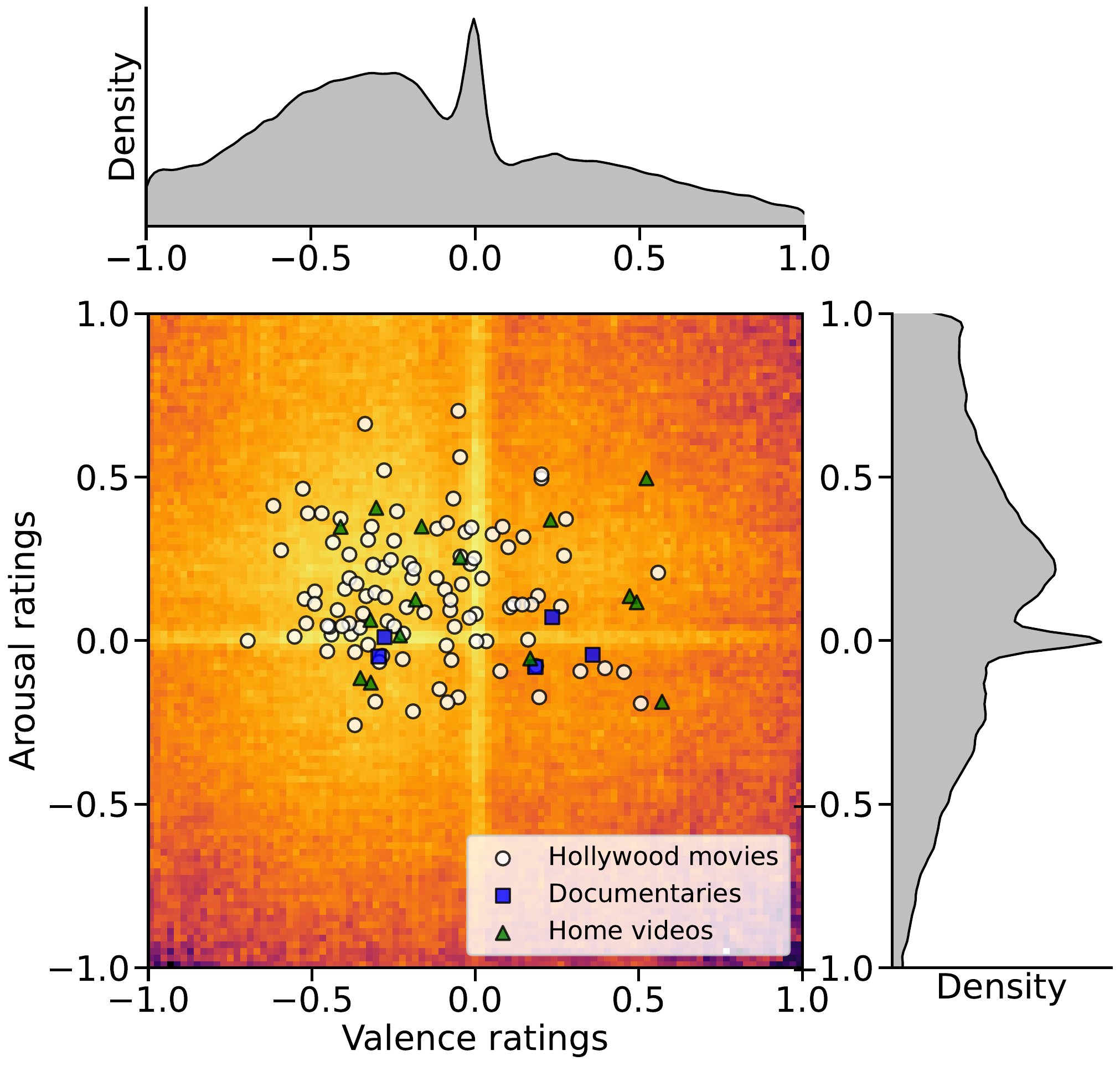}
    \caption{Distribution of valence and arousal ratings across participants. Individual white dots represent the average valence and arousal of the continuous ratings for each video clip for Hollywood movies. Blue squares and green triangles represent the average valence and arousal for documentaries and home videos, respectively. Ratings were binned into 0.02 intervals and the total number of data points was counted within each bin.} 
  
    \label{fig:density_plot}
\end{figure}

\begin{figure}[th]
    \centering
    \includegraphics[width=0.25\textwidth]{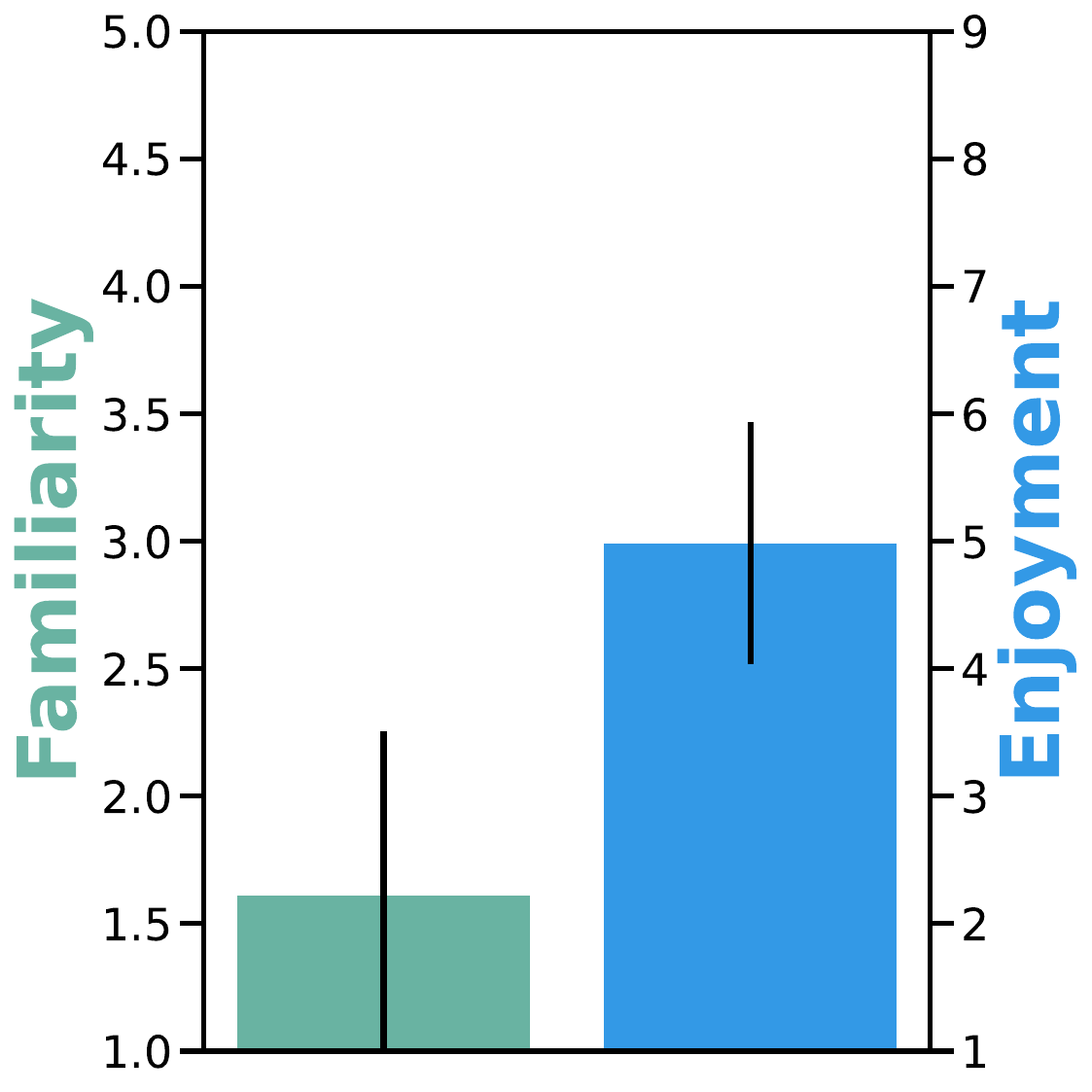}
    \caption{Familiarity and enjoyment ratings across videos for video IDs 0-97.  Vertical black lines indicate 1 SD. }
    
    \label{fig:fe_single_new}
\end{figure}

The distribution of the valence and arousal ratings across all of our videos is shown in Figure~\ref{fig:density_plot}. We found that individual participant ratings were distributed fully across both valence and arousal dimensions which highlights the diversity of the VEATIC dataset. We also collected familiarity and enjoyment ratings for each video across participants (shown in Figure~\ref{fig:fe_single_new}). We found that observers were unfamiliar with the videos used in the dataset as the average familiarity rating was 1.61 for video IDs 0-97. Additionally, observers rated their enjoyment while watching the videos as an average of 4.98 for video IDs 0-97 indicating that observers moderately enjoyed watching and annotating the video clips. Familiarity and enjoyment ratings were not collected for video IDs 98-123 as the annotations for these videos were collected at an earlier time point during data collection which did not include these ratings.

Table~\ref{table:summary} below summarizes the basic statistics of the VEATIC dataset. In a nutshell, VEATIC has a long total video clip duration and a variety of video sources that cover a wide range of contexts and emotional conditions. Moreover, compared to previous datasets, we recruited far more participants to annotate the ratings.

\begin{table}[h]
\footnotesize
\centering
\begin{tabular}{|c|c|}
\hline
\textbf{Attribute} & \textbf{Description} \\ \hline
No. of Frames & $257,601$ \\ \hline % Recalculate this as well 
No. of Videos & $124$ \\ \hline % With Mandy's extra 28 videos
Total No. of Annotators & $192$ \\ \hline
Avg. No. of Annotators per video & $60$ \\ \hline
Length of Videos & 10 s - 2 min 37s \\ \hline
Mean Image Resolution & $854\times480$ \\ \hline
Hollywood movies & $104$ \\ \hline
Documentaries & $5$ \\ \hline
Home videos & $15$ \\ \hline
\end{tabular}
\newline
\caption{Statistics of VEATIC Dataset.}
\label{table:summary}
\end{table}

\section{Experiments}
\begin{figure}[th]
    \centering
    \includegraphics[width=0.5\textwidth]{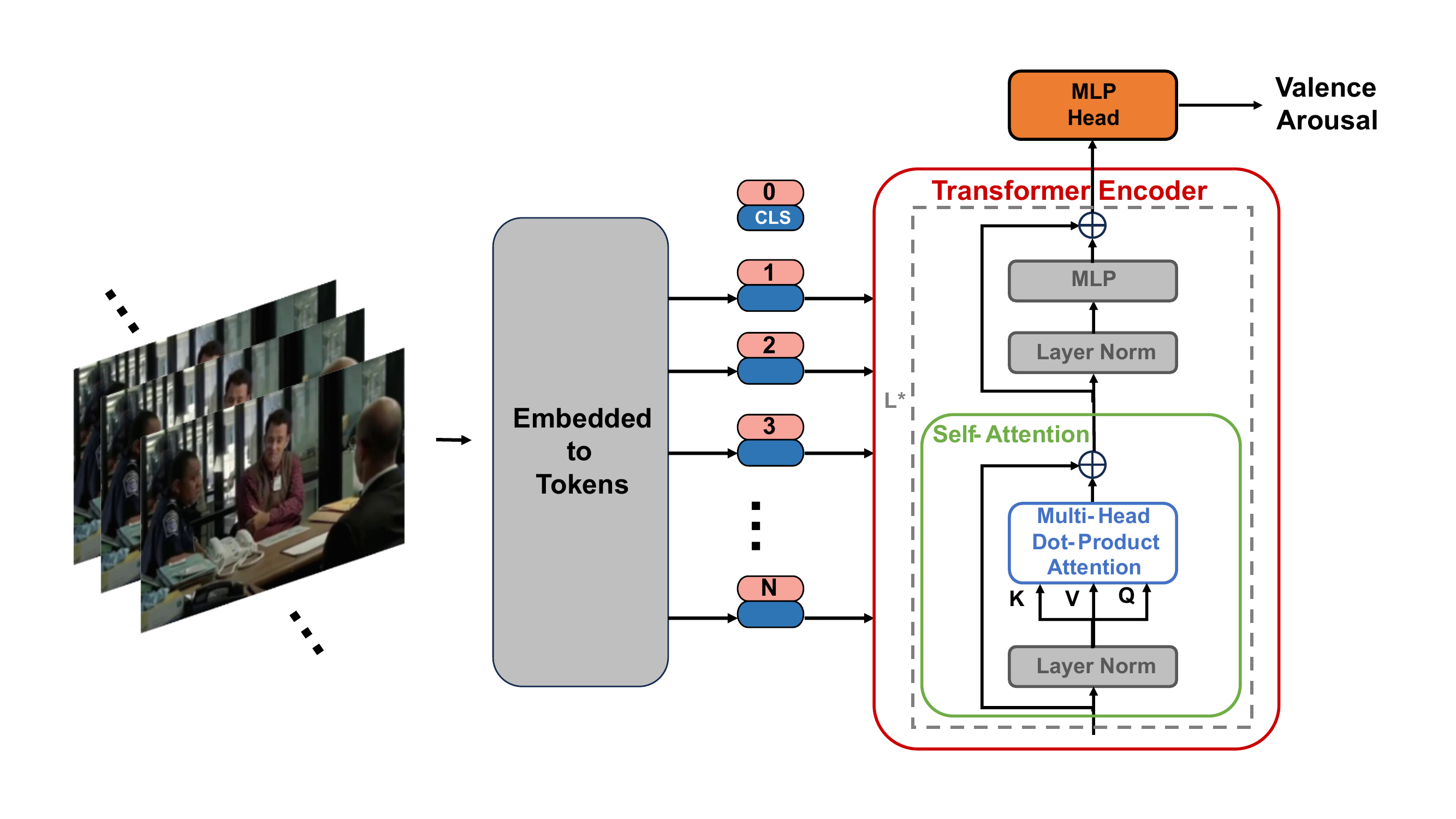}
    \caption{The architecture of the benchmark model for emotion and affect tracking in context task. The model consists of a CNN feature extraction module and a visual transformer for combining temporal information of consecutive frames.}
    \label{fig:Architecture}
\end{figure}

In this study, we propose a new emotion recognition in context task, i.e. to infer the valence and arousal of the selected character via both context and character information in each video frame. Here, we propose a simple baseline model to benchmark the new emotion recognition in context task. The pipeline of the model is shown in Figure~\ref{fig:Architecture}. We adopted two simple submodules: a convolutional neural network (CNN) module for feature extraction and a visual transformer module for temporal information processing. The CNN module structure is adopted from Resnet50~\cite{he2016deep}. Unlike CAER~\cite{lee2019context} and EMOTIC~\cite{kosti2019context}, where facial/character and context features are extracted separately and merged later, we directly encode the fully informed frame. For a single prediction, consecutive $N$ video frames are encoded independently. Then, the feature vectors of consecutive frames are first position embedded and fed into the transformer encoder containing $L$ sets of attention modules. At last, the prediction of arousal and valence is accomplished by a multilayer perceptron (MLP) head.

\vspace{-0.1cm}
\subsection{Loss Function and Training Setup}

The loss function of our baseline model is a weighted combination of two separate losses. The MSE loss regularizes the local alignment of the ground truth of ratings and the model predictions. In order to guarantee the alignment of the ratings and predictions on a larger scale, such as learning the temporal statistics of the emotional ratings, we also utilize the concordance correlation coefficient (CCC) as a regularization. This coefficient is defined as follows,

\begin{equation}
    \rho_c = \frac{2s_{xy}}{s_x^2 + s_y^2 + (\Bar{x} - \Bar{y})^2}
    \label{equ:ccc}
\end{equation}

where $s_x$ and $s_y$ are the variances of the ground truth and predicted values in a training batch respectively, $\Bar{x}$ and $\Bar{y}$ are the corresponding mean values, and $s_{xy}$ is the respective covariance value. After computing the CCC of both arousal and valence, the CCC loss is computed as follows,

\begin{equation}
    \mathcal{L}_{CCC} = 1 - \frac{\rho_a + \rho_v}{2}
\end{equation}

where $\rho_a$ and $\rho_v$ are the concordance correlation coefficient (CCC) for the arousal and valence, respectively. Together, our final training loss is defined as,

\begin{equation}
   \mathcal{L} = \mathcal{L}_{CCC} + \lambda\mathcal{L}_{MSE}
\end{equation}

During training, $\lambda$ is set to $0.1$. The sliding window size $N$ is $5$ and the depth of the visual transformer $L$ is set to $6$. We train our emotion and affect tracking baseline model end-to-end with Adam optimizer~\cite{kingma2014adam}, where $\beta_1=0.9$ and $\beta_2=0.999$. The learning rate is set to $0.0005$ with a cosine annealing schedule. The first CNN feature extraction module is initialized using a pretrained Resnet50~\cite{he2016deep} on Imagenet~\cite{russakovsky2015imagenet} while the visual transformer is initialized using Kaiming initialization~\cite{he2015delving}. The batch size is set to $20$, and we utilize $4$ GPUs for training. For each video, the first 70\% of the frames are for training and the rest 30\% of the frames are for testing.

\begin{table*}[th]
\footnotesize
\begin{center}
\begin{tabular}{*{10}{c}}
  \toprule
  \multirow{2}*{\textbf{Frame Type}} & \multicolumn{4}{c}{\textbf{Valence}} & \multicolumn{4}{c}{\textbf{Arousal}} \\
  
  \cmidrule(lr){2-5}\cmidrule(lr){6-9}
  
  & \textbf{CCC$\uparrow$} & \textbf{PCC$\uparrow$} & \textbf{RMSE$\downarrow$} & \textbf{SAGR$\uparrow$} & \textbf{CCC$\uparrow$} & \textbf{PCC$\uparrow$} & \textbf{RMSE$\downarrow$} & \textbf{SAGR$\uparrow$} \\
  \midrule
  \textbf{Fully Informed} & \textbf{0.6678} & \textbf{0.6967} & \textbf{0.3084} & \textbf{0.8149} & \textbf{0.6296} & \textbf{0.6584} & \textbf{0.2410} & 0.7637 \\
  
  \textbf{Character Only} & 0.5116 & 0.5609 & 0.3776 & 0.7451 & 0.5725 & 0.6247 & 0.2333 & 0.7497 \\
  
  \textbf{Context Only} & 0.6185 & 0.6567 & 0.3245 & 0.8071 & 0.6088 & 0.6181 & 0.2416 & \textbf{0.7828} \\
  \bottomrule
\end{tabular}
\end{center}
\caption{Performance of our proposed model on fully-informed, character-only, and context-only conditions. Inference via both character and context information, the model performs the best. It shows the importance of both context and character information in emotion and affect tracking tasks.}
\label{table:ViT}
\end{table*}

\begin{table}[th]
\resizebox{\columnwidth}{!}{
\begin{tabular}{*{6}{c}}
  \toprule
  \multirow{2}*{\textbf{Method}} & \multicolumn{3}{c}{\textbf{RMSE$\downarrow$}} & \multirow{2}*{\textbf{Method}} & \multirow{2}*{\textbf{ACC$\uparrow$}}\\
  
  \cmidrule(lr){2-4}
  
  & \textbf{Valence} & \textbf{Arousal} & \textbf{Overall}\\
   \midrule
  \textbf{EMOTIC}  &1.1730  &1.2900 &1.2315 & \textbf{CAER-NET-S}  & 0.7351\\
    \textbf{Ours} & 1.2151 & 1.3213 & 1.2682 &  \textbf{Ours} & 0.6904\\

  \bottomrule\\
\end{tabular}
}
\caption{Comparison of our fine-tuned proposed method with EMOTIC and CARE-S pretrained model on their Datasets. Our simple model achieves competitive results, indicating the generalizability of VEATIC.}
\label{table:ModelComparison}
\end{table}

\subsection{Evaluation Metrics}
When testing our model's performance, we utilize the concordance correlation coefficient (CCC, $\rho_c$)  as one of the evaluation metrics. In addition to CCC, we also utilize the Pearson correlation coefficient (PCC), the root mean square error (RMSE), and the sign agreement (SAGR) to evaluate the model's performance. The SAGR metric is defined as follows.

\begin{equation}
    \textit{SAGR}(X, Y) = \frac{1}{N}\sum_{i=1}^{N}\delta(\textit{sign}(x_i),\textit{sign}(y_i))
\end{equation}

where $\bar{x}$ and $\bar{y}$ are the means of sample X and Y, and $\delta(x, y)$ denotes the Kronecker delta $\delta(x, y) = 1$ if $x = y$; Otherwise $\delta(x, y) = 0$.

The SAGR measures how much the signs of the individual values of two vectors X and Y match. It takes on values in [0, 1], where 1 represents the complete agreement and 0 represents a complete contradiction. The SAGR metric can capture additional performance information than others. For example, given a valence ground truth of 0.2, predictions of 0.7 and -0.3 will lead to the same RMSE value. But clearly, 0.7 is better suited because it is a positive valence.

\subsection{Benchmark Results}
We benchmark the new emotion recognition in context task using the aforementioned 4 metrics, CCC, PCC, RMSE, and SAGR. Results are shown in Table~\ref{table:ViT}. Compared to other datasets, our proposed simple method is on par with state-of-the-art methods on their datasets.

We also investigate the importance of context and character information in emotion recognition tasks by feeding the context-only and character-only frames into the pretrained model on fully-informed frames. In order to obtain fair comparisons and exclude the influence of frame pixel distribution differences, we also fine-tune the pretrained model on the context-only and character-only frames. The corresponding results are shown in Table~\ref{table:ViT} as well. Without full information, the model performances drop for both context-only and character-only conditions.

In order to show the effectiveness of the VEATIC dataset, we utilized our pretrained model on VEATIC, fine-tuned it on other datasets, and tested its performance. We only tested for EMOTIC~\cite{kosti2019context} and CAER-S~\cite{lee2019context} given the simplicity of our model and the similarity of our model to the models proposed in other dataset papers. The results are shown in Table~\ref{table:ModelComparison}. Our pretrained model performs on par with the proposed methods in EMOTIC~\cite{kosti2019context} and CAER-S~\cite{lee2019context}. Thus, it shows the effectiveness of our proposed VEATIC dataset.

\section{Discussion}
Understanding how humans infer the emotions of others is essential for researchers understanding of social cognition. While psychophysicists conduct experiments, they need specific stimulus sets to design experiments. However, among published datasets, there is currently no context-based video dataset that contains continuous valence and arousal ratings. The lack of this kind of datasets also prevents researchers from developing computer vision algorithms for the corresponding tasks. Our proposed VEATIC dataset fills in this important gap in the field of computer vision and will be beneficial for psychophysical studies in understanding emotion recognition.

During data collection, participants continuously tracked and rated the emotions of target characters in the video clips which is different from general psychophysical experiments where responses are collected after a delay. This design in our dataset was vital in order to mimic the real-time emotion processing that occurs when humans process emotions in their everyday lives. Additionally, emotion processing is not an immediate process and it relies heavily on the temporal accumulation of information over time in order to make accurate inferences about the emotions of others.

The strength of the VEATIC dataset is that it mimics how humans perceive emotions in the real world: continuously and in the presence of contextual information both in the temporal and spatial domain. Such a rich dataset is vital for future computer vision models and can push the boundaries of what current models can accomplish. With the creation of more rich datasets like VEATIC, it may be possible for future computer vision models to perceive emotions in real-time while interacting with humans.

\section{Conclusion}
In this study, we proposed the first context based large video dataset, \textbf{VEATIC}, for continuous valence and arousal prediction. Various visualizations show the diversity of our dataset and the consistency of our annotations. We also proposed a simple baseline algorithm to solve this challenge. Empirical results prove the effectiveness of our proposed method and the VEATIC dataset.

%%%%%%%%%%%%%%%%%%%%%%%%%%%%%%%%%%%%%%%SUPPLEMENTARY%%%%%%%%%%%%%%%%%%%%%%%%%%%%%%%%%%%%%%%%%%%%%%%%%%%%%
\clearpage

\title{Supplementary for VEATIC: Video-based Emotion and Affect Tracking in Context Dataset}

% \author{Zhihang Ren*\quad Jefferson Ortega*\quad Yifan Wang*\quad Zhimin Chen\quad David Whitney\\
% University of California, Berkeley\\
% {\tt\small \{peter.zhren,jefferson\_ortega,wyf020803,zhimin,dwhitney\}@berkeley.edu}
% \and
% Yunhui Guo\\
% University of Texas at Dallas\\
% {\tt\small yunhui.guo@utdallas.edu}
% \and
% Stella Yu\\
% University of Michigan, Ann Arbor\\
% {\tt\small stellayu@umich.edu}
% }

\author{Zhihang Ren*$^{1}$, Jefferson Ortega*$^{1}$, Yifan Wang*$^{1}$, Zhimin Chen$^{1}$, Yunhui Guo$^{2}$,\\ Stella X. Yu$^{1,3}$, David Whitney$^{1}$\\
$^{1}$University of California, Berkeley, $^{2}$University of Texas at Dallas,\\ $^{3}$University of Michigan, Ann Arbor\\
$^{1}${\tt\small \{peter.zhren,jefferson\_ortega,wyf020803,zhimin,dwhitney\}@berkeley.edu},\\
$^{2}${\tt\small yunhui.guo@utdallas.edu},$^{3}${\tt\small stellayu@umich.edu}
}

\maketitle
\thispagestyle{empty}

\def\thefootnote{*}\footnotetext{These authors contributed equally to this work.}

%%%%%%%%% BODY TEXT

\section{More About Stimuli}

All videos used in the VEATIC dataset were selected from an online video-sharing website (YouTube). The VEATIC dataset contains 124 video clips, 104 clips from Hollywood movies, 15 clips from home videos, and 5 clips from documentaries or reality TV shows. Specifically, we classify Documentary videos as any videos that show candid social interactions but have some form of video editing, while home videos refer to videos that show candid social interactions without any video editing. All Videos in the dataset had a frame rate of 25 frames per second and ranged in resolution with the lowest being 202 x 360 and the highest being 1920 x 1080. 

\begin{figure*}[th]
    \centering
    \includegraphics[width=0.82\textwidth]{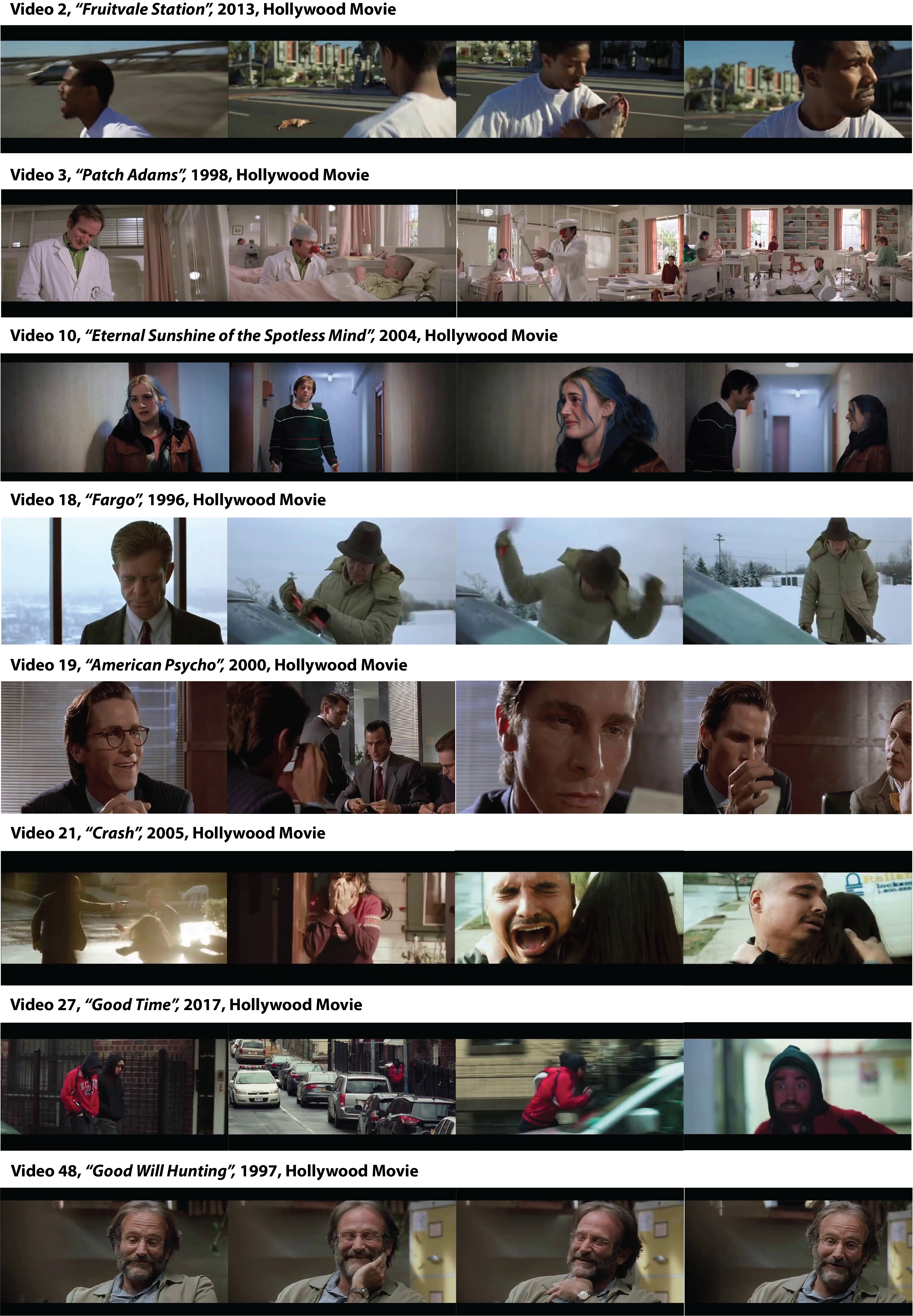}
    \caption{More sample video frames in VEATIC. The video clips in VEATIC contain various backgrounds, lighting conditions, character interactions, etc., making it a comprehensive dataset for not only emotion recognition tasks but also other video understanding tasks.}
    \label{fig:Overview1}
\end{figure*}

\begin{figure*}[th]
    \centering
    \includegraphics[width=0.8\textwidth]{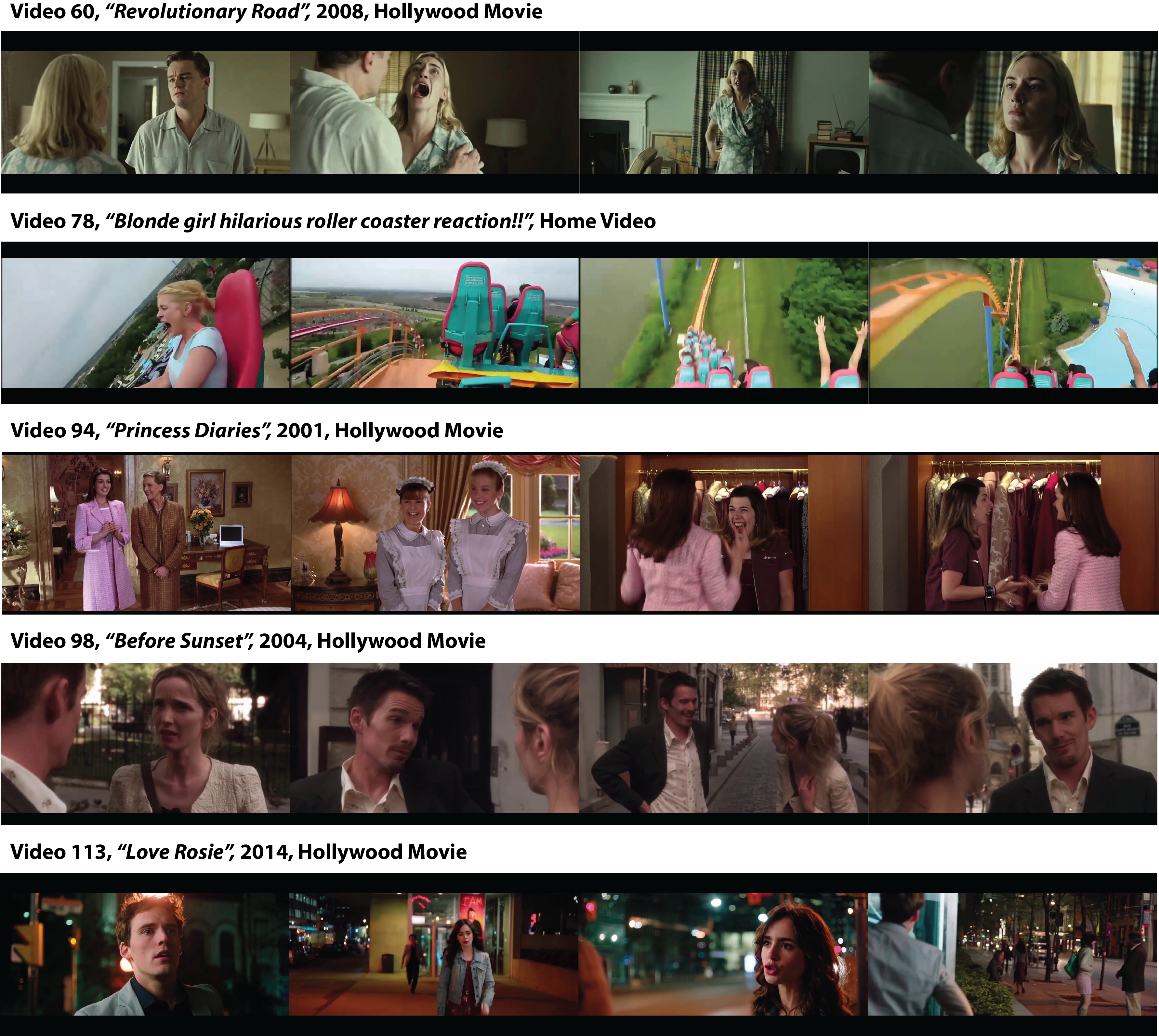}
    \caption{Sample video frames of unselected characters and pure background in VEATIC. The first sample frame in each row shows the selected character. The rest sample frames are either unselected characters or pure backgrounds.}
    \label{fig:Overview2}
\end{figure*}

Except for the overview of video frames in Figure 2, we show more samples in Figure~\ref{fig:Overview1}. Moreover, unlike previously published datasets where most frames contain the main character~\cite{kossaifi2017afew,kollias2018aff,kosti2019context}, VEATIC not only has frames containing the selected character but also there are lots of frames containing unselected characters and pure backgrounds (Figure~\ref{fig:Overview2}). Therefore, VEATIC is more similar to our daily life scenarios, and the algorithms trained on it will be more promising for daily applications.

\vspace{-0.1cm}
\section{Annotation Details}

In total, we had 192 participants who annotated the videos in the VEATIC dataset. Eighty-four participants annotated video IDs 0-82. One hundred and eight participants annotated video IDs 83-123 prior to the planning of the VEATIC dataset. In particular, Fifty-one participants annotated video IDs 83-94, twenty-five participants annotated video IDs 95-97, and 32 participants annotated video IDs 98-123. 

\begin{figure*}[th]
    \centering
    \includegraphics[width=0.9\textwidth]{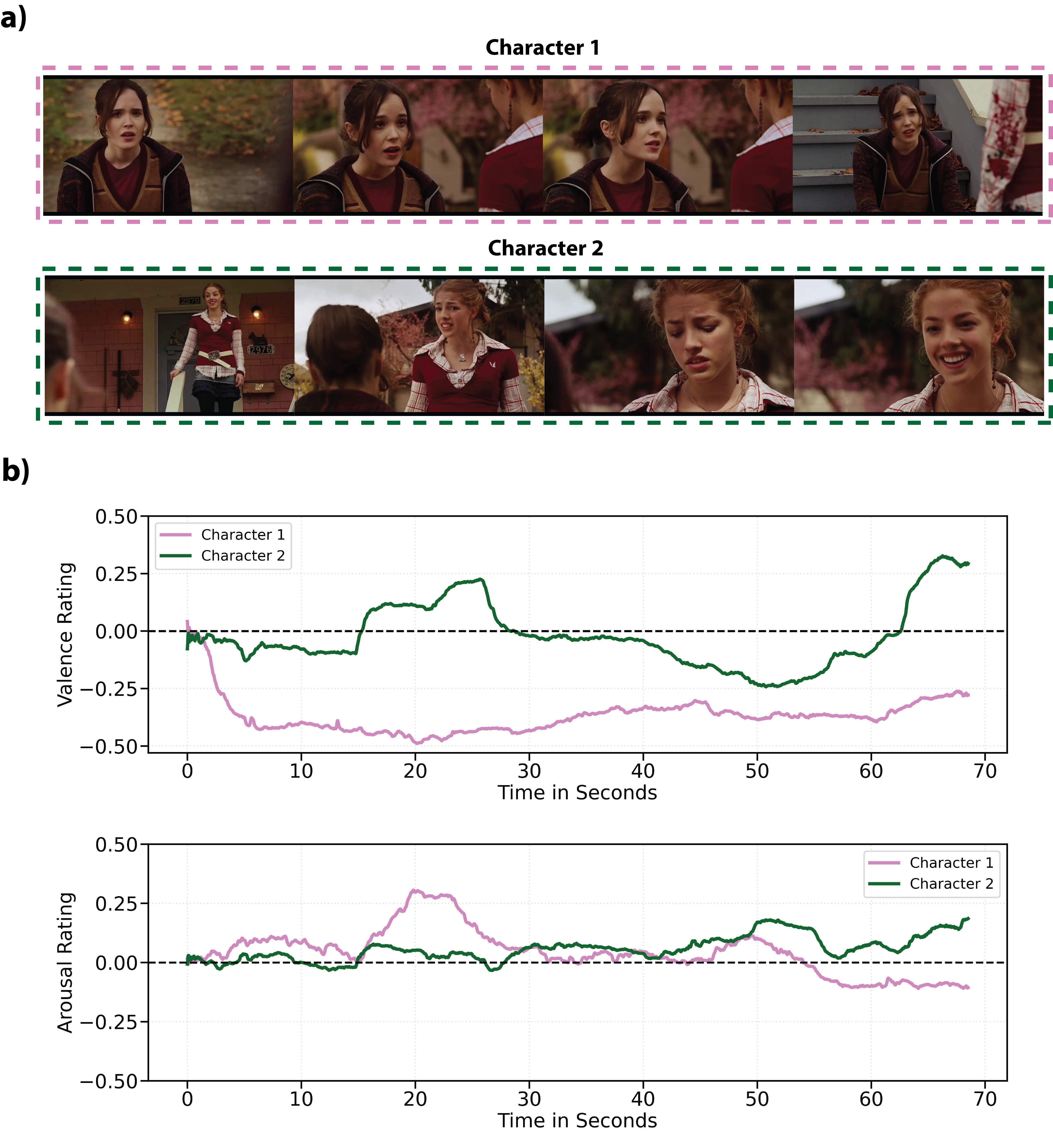}
    \caption{Example of different ratings of the same video in VEATIC. (a). The two selected characters. (b). The continuous emotion ratings of corresponding characters. The same color indicates the same character. A good emotion recognition algorithm should infer the emotion of two characters correspondingly given the interactions between characters and the exact same context information.}
    \label{fig:doublerating}
\end{figure*}

Another novelty of the VEATIC dataset is that it contains videos with interacting characters and ratings for separate characters in the same video. These videos are those with video IDs 98-123. For each consecutive video pair, the video frames are exactly the same, but the continuous emotion ratings are annotated based on different selected characters. Figure~\ref{fig:doublerating} shows an example. In this study, we first propose this annotation process because it affords future algorithms a way to test whether models learn the emotion of the selected characters given the interactions between characters and the exact same context information. A good emotion recognition algorithm should deal with this complicated situation.

\vspace{-0.2cm}
\section{Outlier Processing}

We assessed whether there were any noisy annotators in our dataset by computing each individual annotator's agreement with the consensus. This was done by calculating the Pearson correlation between each annotator and the leave-one-out consensus (aggregate of responses except for the current annotator) for each video. Only one observer in our dataset had a correlation smaller than .2 with the leave-one-out consensus rating across videos. We chose .2 as a threshold because it is often used as an indicator of a weak correlation in psychological research. Importantly, if we compare the correlations between the consensus of each video and a consensus that removes the one annotator who shows weak agreement, we get a very high correlation (r = 0.999) indicating that leaving out that subject does not significantly influence the consensus response in our dataset. Thus, we decided to keep the annotator with weak agreement in the dataset in order to avoid removing any important alternative annotations to the videos.

\section{Subject Agreement Across Videos}

\begin{figure*}[th]
    \centering
    \includegraphics[width=0.95\textwidth]{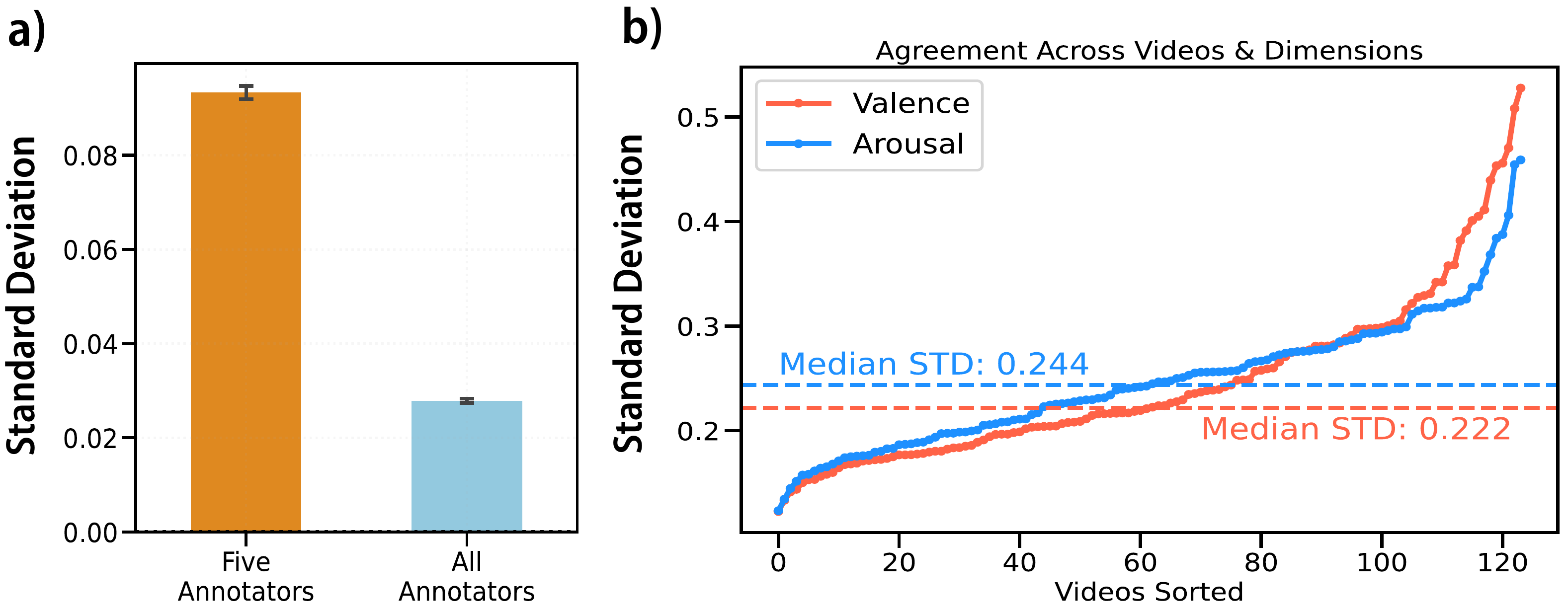}
    \caption{a) Five annotators' response standard deviation vs. all annotators' response standard deviation. Testing a small number of annotators can lead to substantial imprecision in annotations. Increasing the number of annotators, as in this study, greatly improves precision. b) Annotators' response standard deviation for each video. Red and blue solid lines indicate the standard deviation of annotators' responses for valence and arousal, in each video, respectively. The results are sorted based on the standard deviation of each video for visualization purposes. The dashed lines show the median standard deviation for each dimension. The mean values for standard deviations of valence and arousal are the same with $\mu=0.248$.}
    \label{fig:n_variance}
\end{figure*}

A benefit of the VEATIC dataset is that it has multiple annotators for each video with the minimum number of annotators for any given video being 25 and the maximum being 73. Emotion perception is subjective and observers judgments can vary across multiple people. Many of the previously published emotion datasets have a very low number of annotators, often having only single digit \begin{math}(n < 10)\end{math} number of annotators. Having so few annotators is problematic because of the increased variance across observers. To show this, we calculated how the average rating for each video in our dataset varied if we randomly sampled, with replacement, five versus all annotators. We repeated this process 1000 times for each video and calculated the standard deviation of the recalculated average rating. Figure~\ref{fig:n_variance}a shows how the standard deviation of the consensus rating across videos varies if we use either five or all annotators for each video. This analysis shows that having more annotators leads to much smaller standard deviations in the consensus rating which can lead to more accurate representations of the ground truth emotion in the videos.

Additionally, We investigated how observers' responses varied across videos by calculating the standard deviation across observers for each video. Figure~\ref{fig:n_variance}b shows the standard deviations across videos. We find that the standard deviations for both valence and arousal dimensions were small with valence having an average standard deviation of $\mu=0.248$ and a median of $0.222$ and arousal having an average standard deviation of $\mu=0.248$ and a median of $0.244$, which are comparable with the valence and arousal rating variance from EMOTIC~\cite{kosti2019context}.

\section{Familiraity and Enjoyment Ratings}

\begin{figure*}[th]
    \centering
    \includegraphics[width=0.95\textwidth]{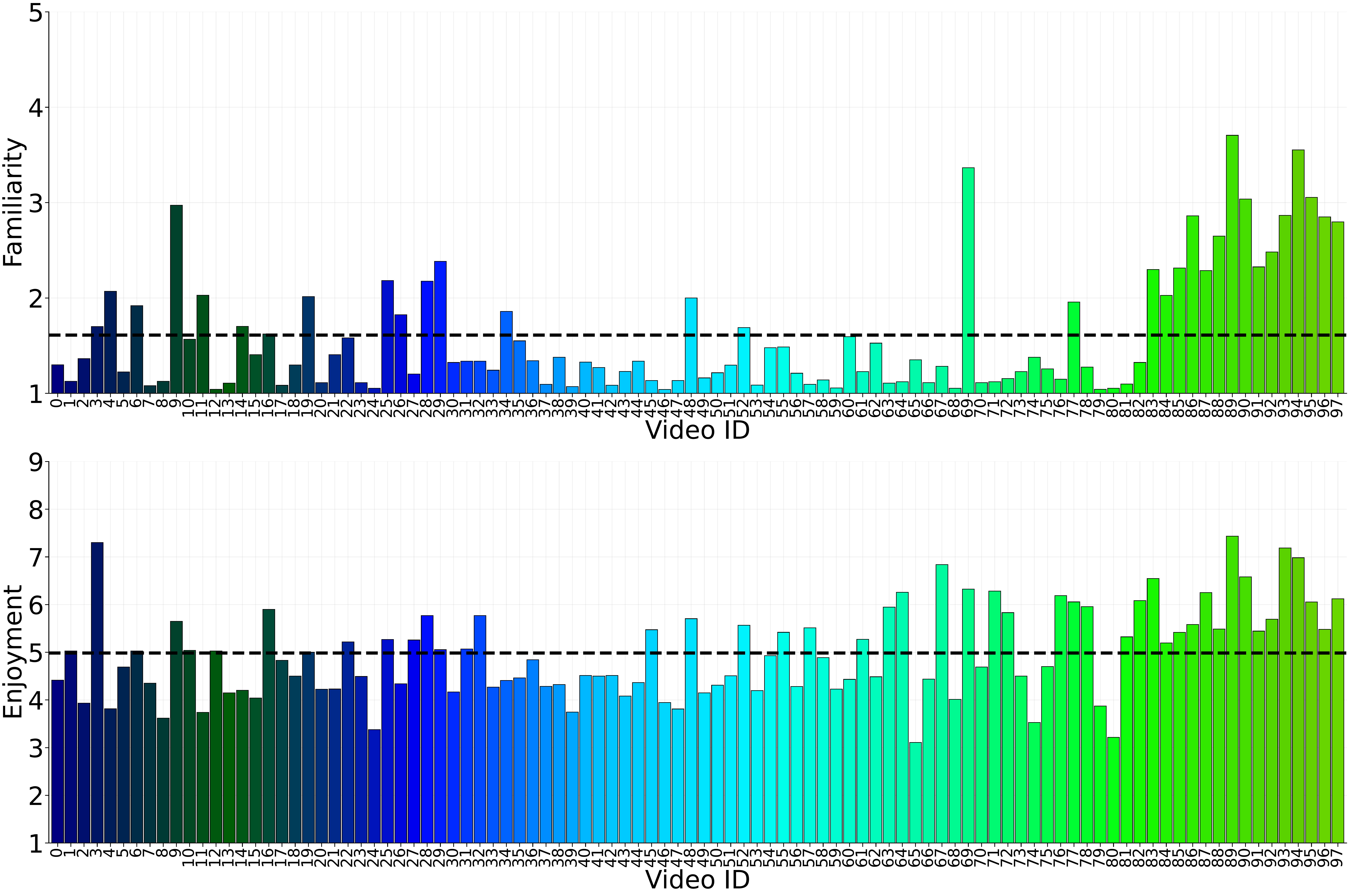}
    \caption{Familiarity and enjoyment ratings across all videos. Each bar represents the average familiarity or enjoyment rating reported by all participants who annotated the video. The average rating across all videos is depicted by the horizontal dashed line in both figures. Video IDs are shown on the x-axis.}
    \label{fig:fam_enjoy}
\end{figure*}

Familiarity and enjoyment ratings were collected for each video across participants, as shown in Figure~\ref{fig:fam_enjoy}. Familiarity and enjoyment ratings for video IDs 0-83 were collected in a scale of 1-5 and 1-9, respectively. Familiarity and enjoyment ratings for video IDs 83-123 were collected prior to the planning of the VEATIC dataset and were collected on a different scale. Familiarity and enjoyment ratings for video IDs 83-97 were collected on a scale of 0-5 and familiarity/enjoyment ratings were not collected for video IDs 98-123. For analysis and visualization purposes, we rescaled the familiarity and enjoyment ratings for video IDs 83-97 to 1-5 and 1-9, respectively, to match video IDs 0-83. To rescale the familiarity values from 0-5 to 1-5 we performed a linear transformation, we first normalized the data between 0 and 1, then we multiplied the values by 4 and added 1. We rescaled the enjoyment values from 0-5 to 1-9 similarly by first normalizing the data between 0 and 1, then we multiplied the values by 8 and added 1. As a result, the average familiarity rating was 1.61 while the average enjoyment rating was 4.98 for video IDs 0-97. 

%%%%%%%%%%%%%%%%%%%%%%%%%%%%%%%%%%%%%%%SUPPLEMENTARY%%%%%%%%%%%%%%%%%%%%%%%%%%%%%%%%%%%%%%%%%%%%%%%%%%%%%

{\small
\bibliographystyle{ieee_fullname}
\bibliography{egbib}
}

\end{document}